\documentclass[10pt,twocolumn,letterpaper]{article}

\usepackage[pagenumbers]{cvpr} 

\usepackage{graphicx}
\usepackage{amsmath}
\usepackage{amssymb}
\usepackage{booktabs}

\usepackage{xcolor}
\definecolor{citecolor}{RGB}{34,139,34} 
\usepackage[pagebackref,breaklinks,colorlinks,
citecolor=citecolor,bookmarks=false]{hyperref}

\usepackage{makecell}
\usepackage{tabulary}
\usepackage{multirow}

\usepackage[capitalize]{cleveref}
\crefname{section}{Sec.}{Secs.}
\Crefname{section}{Section}{Sections}
\Crefname{table}{Table}{Tables}
\crefname{table}{Tab.}{Tabs.}

\definecolor{bittersweet}{rgb}{1.0, 0.44, 0.37}
\definecolor{mygreen}{rgb}{0.29, 0.7, 0.48}

\def\ModelName{\textsc{LoopITR}}

\newcommand{\tabincell}[2]{\begin{tabular}{@{}#1@{}}#2\end{tabular}}

\usepackage[font=small,labelfont=bf]{caption}  
\usepackage{makecell}
\usepackage{tabulary}
\definecolor{demphcolor}{RGB}{144,144,144}
\newcommand{\demph}[1]{\textcolor{demphcolor}{#1}}
\definecolor{mygray}{gray}{0.4}
\usepackage{pifont}
\newcommand{\cmark}{\color{mygray}\ding{51}}%

\newlength\savewidth\newcommand\shline{\noalign{\global\savewidth\arrayrulewidth
  \global\arrayrulewidth 1pt}\hline\noalign{\global\arrayrulewidth\savewidth}}

\newcommand{\tablestyle}[2]{\setlength{\tabcolsep}{#1}\renewcommand{\arraystretch}{#2}\centering\footnotesize}
\makeatletter\renewcommand\paragraph{\@startsection{paragraph}{4}{\z@}
  {.5em \@plus1ex \@minus.2ex}{-.5em}{\normalfont\normalsize\bfseries}}\makeatother
\def\x{$\times$} 

\newcolumntype{C}[1]{>{\centering\arraybackslash}p{#1}}
\newcolumntype{R}[1]{>{\raggedleft\arraybackslash}p{#1}}
\newcolumntype{L}[1]{>{\raggedright\arraybackslash}p{#1}}

\usepackage{listings}
\definecolor{codeblue}{rgb}{0.25,0.5,0.5}
\definecolor{codekw}{rgb}{0.85, 0.18, 0.50}
\def\confName{CVPR}
\def\confYear{2022}

\begin{document}

\title{
\ModelName: Combining Dual and Cross Encoder Architectures for Image-Text Retrieval
}

\author{Jie Lei$^1$, Xinlei Chen$^2$, Ning Zhang$^2$, Mengjiao Wang$^2$, \\Mohit Bansal$^1$, Tamara L. Berg$^2$, Licheng Yu$^2$\\ 
  $^1$UNC Chapel Hill \quad $^2$Meta AI \\
  \texttt{\small\{jielei, mbansal\}@cs.unc.edu} \\
  \texttt{\small\{xinleic, ningzhang, mengjiaow, tlberg, lichengyu\}@fb.com}
  }

\maketitle

\begin{abstract}
Dual encoders and cross encoders have been widely used for image-text retrieval.
Between the two, the dual encoder encodes the image and text independently followed by a dot product, while the cross encoder jointly feeds image and text as the input and performs dense multi-modal fusion.
These two architectures are typically modeled separately without interaction. 
In this work, we propose \ModelName, which combines them in the same network for joint learning.
Specifically, we let the dual encoder provide hard negatives to the cross encoder, and use the more discriminative cross encoder to distill its predictions back to the dual encoder.
Both steps are efficiently performed together in the same model.
Our work centers on empirical analyses of this combined architecture, putting the main focus on the design of the distillation objective.
Our experimental results highlight the benefits of training the two encoders in the same network, and demonstrate that distillation can be quite effective with just a few hard negative examples.
Experiments on two standard datasets (Flickr30K and COCO) show our approach achieves state-of-the-art dual encoder performance when compared with approaches using a similar amount of data. 
\end{abstract}

\section{Introduction} \label{sec:intro}
There are two widely used architectures for image-text retrieval: $(i)$ \textit{dual encoder}, 
with two separate streams (weights might be shared~\cite{geigle2021retrieve,li2020unimo,gur-etal-2021-cross-modal}) that encodes the image and text modalities independently, and the matching score is obtained via a simple dot product~\cite{frome2013devise,chen2021learning,radford2021learning,jia2021scaling}; and $(ii)$ \textit{cross encoder},
which takes the joint sequence of image and text as input, and performs dense cross-modal interactions (e.g., cross-attention~\cite{vaswani2017attention}) in a single stream. 
The final image-text matching score is often predicted with a classifier~\cite{chen2019uniter,li2019visualbert,li2020oscar}. 
These two types of encoders are typically learned separately without interaction.

\textbf{}
\begin{figure}[!t]
  \centering
  \includegraphics[width=0.94\linewidth]{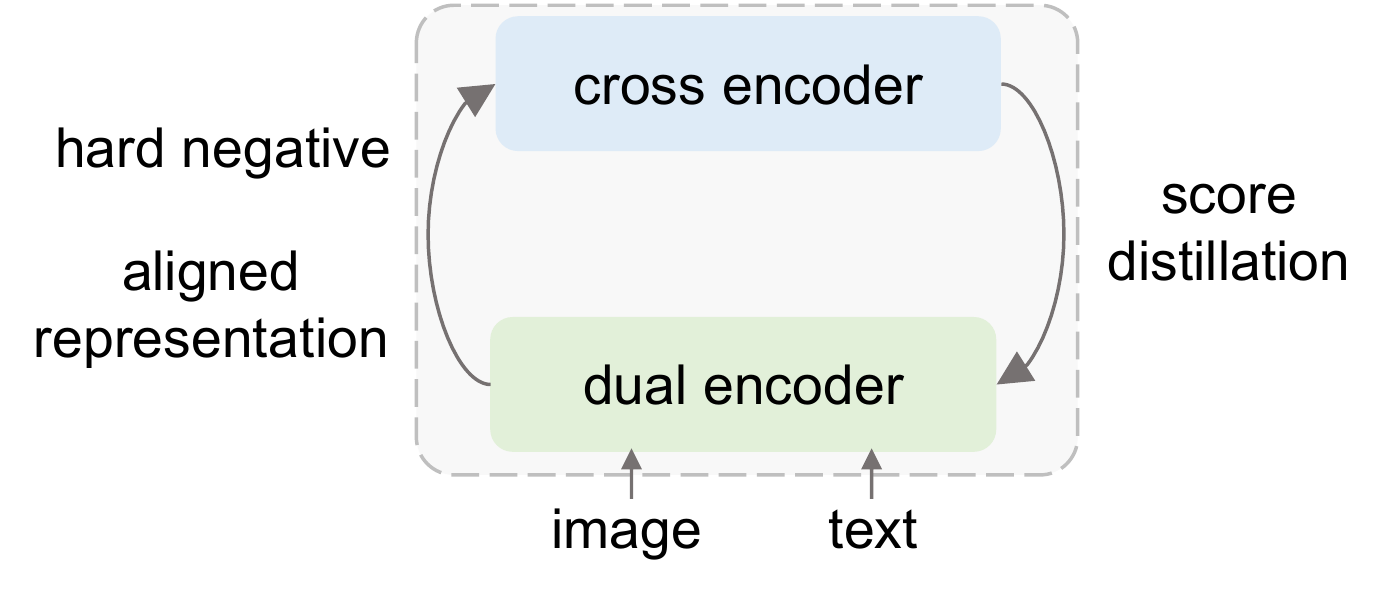}
  \caption{\textbf{\ModelName~architecture} for image-text retrieval. 
Dual encoder and cross encoder architectures are typically learned separately, whereas in \ModelName, we combine them together in the same network via a loop interaction: ($i$) \textit{bottom-up}, dual encoder provides aligned representations and hard negatives for cross encoder learning; ($ii$) \textit{up-down}, cross encoder's scores are used as dual encoder's soft label for additional supervision.
With this loop structure, we observe a performance gain for both encoders.
}
  \label{fig:loop}
\end{figure}

In this work, we propose \ModelName~(pronounced similarly to \textit{Jupiter}), a new model that integrates the dual and cross encoder architectures into a single network via a loop-style structure for image-text retrieval, as in Figure~\ref{fig:loop}. 
Given a pair of image and text, we first feed both inputs into the dual encoder, independently encoding them into their respective embedding sequences, and then apply a cross encoder on top of the two encoded sequences.
Rather than only injecting the supervision on the very top of the model as~\cite{lu2019vilbert,tan2019lxmert,li2021align}, we make both encoders learn to align the image and text modalities.
The dual encoder is trained with a contrastive loss, which has shown to be good at learning representations~\cite{chen2020simple,he2020momentum,chen2021empirical,radford2021learning}. 
Meanwhile, we also use the dual encoder to mine in-batch hard negatives for cross encoder training as~\cite{li2021align}. 
Different from previous work~\cite{chen2019uniter} which requires additional offline computation to mine the hard negatives at every epoch, our approach does not introduce any extra training cost, as the hard negatives are readily available from the dual encoder's contrastive scores. 
On the other hand, the fine-grained alignment information from the cross encoder also benefits the learning of the dual encoder --
more informative signals flow back to the dual encoder via back-propagation.
Empirically, we show the two encoders mutually benefit from joint training.

Cross encoders tend to capture richer information than dual encoders due to their deeper cross-modal interaction. Thus, we further add a distillation loss to the architecture by distilling the image-text matching scores from the cross encoder to the dual encoder.
As distillation for image-text retrieval has rarely been explored, in this work, we put our focus on the designs of this distillation objective and present comprehensive ablations around it.
We first examine how to construct the score distributions for the distillation between the two architectures. 
The standard distillation objective~\cite{hinton2015distilling} is designed for classification tasks and requires a teacher score distribution as input. 
To extend the distillation for image-text retrieval, we follow~\cite{miech2021thinking} to construct a set of negatives for each positive pair. 
In~\cite{miech2021thinking}, all in-batch negatives are used, and only a single distribution for each image-text pair is constructed (from a text to its image negatives).
In our study, we find that distillation can be effective with \textit{a few hard negatives}, and that both image negatives and text negatives are useful. 
These findings lead to a more efficient distillation objective for image-text retrieval.

Our second investigation is on the type of teacher. 
By default, we use an online cross encoder teacher that learns along with the dual encoder student in the same network. 
We compare it with two other designs, including using an offline trained model~\cite{miech2021thinking} and a momentum model~\cite{he2020momentum}. 
We observe all of these teachers work reasonably well, but the simplest online teacher performs the best.
Besides, using an online teacher saves both memory (as no extra model storage is needed) and computation cost (by reusing online computation results from cross encoder training).
This also simplifies the distillation process into a single run, eliminating the need to train a teacher model separately~\cite{hinton2015distilling,miech2021thinking}.

Most existing image-text retrieval methods~\cite{chen2019uniter,li2020oscar} adopt a two-stage pipeline, pre-training followed by fine-tuning. 
Thus, we also study when to perform the distillation in these two stages. 
Experimental results suggest that distillation in pre-training gives more performance gain than that in fine-tuning, and the best performance is achieved by applying distillation in both stages.

Overall, our contributions are: ($i$) \ModelName, a unified architecture that combines dual and cross encoders in the same model, with an efficient online distillation objective that improves the dual encoder for image-text retrieval; ($ii$) We present a comprehensive ablation of \ModelName, with a focus on the cross-to-dual distillation objective design.

\section{Related Work} \label{sec:related}

\paragraph{Language-based Retrieval} 
has been widely studied in various scenarios, e.g., document or passage retrieval~\cite{nogueira2019passage,karpukhin2020dense,nguyen2016ms}, image retrieval~\cite{kiros2014unifying,jia2021scaling,faghri2017vse++,radford2021learning,chen2021learning,chen2019uniter}, video retrieval~\cite{xu2016msr,liu2019use,lei2021less}, and moment retrieval~\cite{anne2017localizing,lei2020tvr,li2020hero}.
Particularly, for image-text retrieval, there are two main-stream methods: 
($i$) \textit{dual encoder}, where the input text queries and the images are encoded separately into dense vectors of the same visual semantic embedding (VSE) space using two encoders~\cite{kiros2014unifying,jia2021scaling,faghri2017vse++,radford2021learning,chen2021learning,miech2021thinking,geigle2021retrieve,gur-etal-2021-cross-modal}. 
This family of models runs fast, as the image and text similarity is simply computed via a dot product between their embeddings;
and ($ii$) \textit{cross encoder}~\cite{chen2019uniter,lu2019vilbert,li2019visualbert,li2020unicoder,li2020oscar,gan2020large,zhang2021vinvl}, where the input text and image pair are jointly encoded by a cross encoder architecture, \eg., a transformer encoder~\cite{vaswani2017attention}. While they are often treated as separate architectures, in this work we study a unified architecture that jointly learns both together.
Our work shares a similar model architecture as~\cite{li2021align}, where both have the dual encoders as the base and a cross encoder on top.
However,~\cite{li2021align} primarily aims to improve the cross encoder, while we lay more focus on utilizing the more powerful cross encoder to improve the fast dual encoder. 
We provide a comprehensive analyses around this design.

\paragraph{Distillation}\cite{bucilua2006model,hinton2015distilling} is a technique for transferring knowledge from a relatively more powerful but complex model (or an ensemble of models) to a simpler one.
This has been widely studied in vision~\cite{ba2013deep,howard2017mobilenets,fang2021seed,touvron2021training}, language ~\cite{kim2016sequence,sanh2019distilbert,jiao2019tinybert,sun2020mobilebert,zhang2021adversarial,izacard2020distilling}, and multi-modal~\cite{wang2020minivlm,fang2021compressing,miech2021thinking} domains.
The students in most of these work learns from an offline trained teacher model.
As comparison, we explore using an online teacher~\cite{zhang2018deep,sun2019deeply} that is trained in the same network together with the student.
Notably most previous works share the same prediction format between the teacher and student (\eg., both are classification scores).
However, in our unified model the cross encoder teacher produces a two-way classification score while the dual encoder student produces a cosine similarity score, causing discrepancy for distillation. 
Thus, we propose an additional simple but important pre-processing step before distillation.

\section{Method} \label{sec:method}
\begin{figure*}[!ht]
  \centering
  \includegraphics[width=\linewidth]{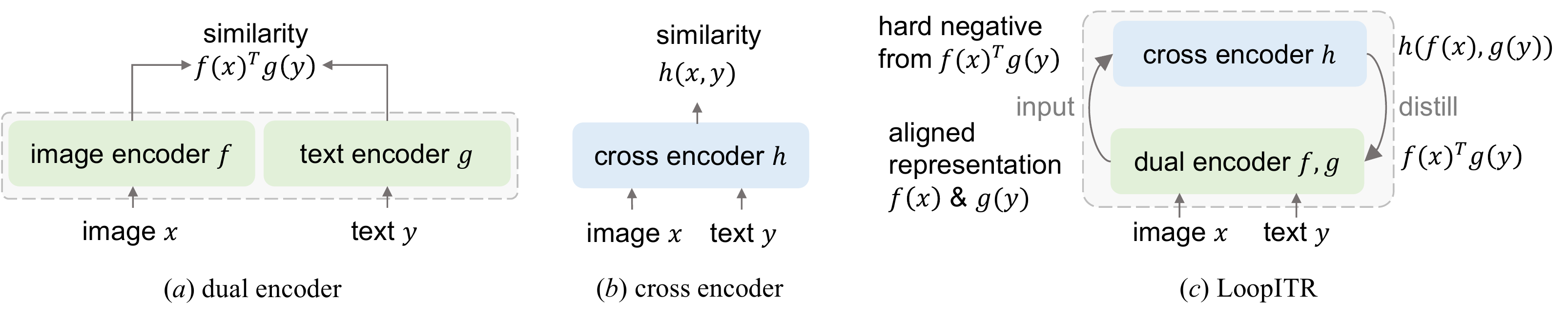}
  \vspace{-10pt}
  \caption{\textbf{Comparison of \ModelName~to cross encoder and dual encoder architectures.} 
  (\textbf{\textit{a}}): dual encoder separately encodes image $x$ and text $y$ into their respective sequence representations $f(x)$ and $g(y)$, and then produces a similarity score as the dot product their pooled representations $\phi(f(x))^T\phi(g(y))$ ($\phi$ is a pooling function, we omit it in the figure for brevity). 
  (\textbf{\textit{b}}): cross encoder $h$ takes image $x$ and text $y$ jointly as inputs, and perform cross-modal interaction (\eg, cross-attention) and then produces a similarity score $h(x, y)$. 
  Cross encoders are typically more powerful than dual encoders as they can perform more fine-grained image-text matching, while dual encoders often run much faster than cross encoders as the image/text representations can be pre-computed, and the similarity scores are computed with a simple dot product.
  (\textbf{\textit{c}}): \ModelName~that combines these two, in which the bottom dual encoder provides aligned representations and hard negatives as inputs the cross encoder, and the cross encoder gives additional supervision to dual encoder via distillation. 
}
\vspace{-5pt}
  \label{fig:comparison_to_dual_cross}
\end{figure*}

In this section, we describe our proposed unified architecture, \ModelName. 
Figure~\ref{fig:comparison_to_dual_cross} shows a conceptual comparison of it to dual and cross encoders.

\paragraph{Dual Encoder} typically consists of two encoders, an image encoder $\mathcal{F}$ that transforms the input image $x$ into a visual feature sequence $\mathcal{F}(x)$, and a text encoder $\mathcal{G}$ that transforms the text sentence $y$ into a textual feature sequence $\mathcal{G}(y)$.
Two projection heads (with pooling) $\phi_{\mathcal{F}}$ and $\phi_{\mathcal{G}}$ are applied to get a vector representation for image and text respectively: $\phi_{\mathcal{F}}(\mathcal{F}(x))$ and $\phi_{\mathcal{G}}(\mathcal{G}(y))$.
The similarity score of image $x$ and text $y$ is measured by dot product:
\begin{align}
    s_{x,y} = \phi_{\mathcal{F}}(\mathcal{F}(x))^T\phi_{\mathcal{G}}(\mathcal{G}(y)).    
\end{align}
For brevity, we omit the notations of $\phi_{\mathcal{F}}$ and $\phi_{\mathcal{G}}$, and denote the similarity as $s_{x,y}=  \mathcal{F}(x)^T\mathcal{G}(y)$.
The goal of the dual encoder is to bring the matched image-text pairs in the embedding space closer than the unmatched ones.
This is often achieved by using a image-text contrastive (ITC) loss:
\begin{align}
    p^{i2t}_{i} {=} \frac{\exp(s_{x_i,y_i} / \tau )}{\sum_{j}\exp(s_{x_i,y_j} / \tau )},\; p^{t2i}_{i} {=} \frac{\exp(s_{x_i,y_i} / \tau )}{\sum_{j}\exp(s_{x_j, y_i} / \tau )}, \label{eq:itc1} \\ 
    \mathcal{L}_{itc} = - \sum_{i=1}^{n} \left( \log  p^{i2t}_{i} + \log p^{t2i}_{i} \right), \quad\quad 
\end{align}
where $\tau$ is a learned temperature parameter.

\paragraph{Cross Encoder.}
While the dual encoder runs fast in retrieval, it only applies an extremely shallow interaction between image and text (dot product), which in turn limits its performance. 
As a comparison, the cross encoder tackles the issue by using multiple layers of cross attention~\cite{vaswani2017attention} for more fine-grained alignment between the input image patches and text tokens.
Specifically, given a cross encoder $\mathcal{H}$, it takes the image $x$ and text $y$ as inputs, and output classification scores (two classes, positive and negative) $\mathbf{h}(x,y) = [h^{pos}_{x,y}; h^{neg}_{x,y}] = \mathcal{H}(x, y) \in \mathbb{R}^2$, where $h^{pos}_{x, y}$ denotes the classification score of the image-text pair $(x, y)$ being the positive pair (in the positive class), and $h^{neg}_{x,y}$ being the negative.
A softmax normalization is then applied $\mathbf{p}_{x,y}^{itm} = \mathrm{softmax}(\mathbf{h}(x,y))$. 
The model is trained using an image-text matching (ITM) objective:
{\small
\begin{align}
    \mathcal{L}_{itm} &{=} \sum_{i=1}^{n}(\text{-}\log\mathbf{p}^{itm}_{x_{i},y_{i}}[0] + \log\mathbf{p}^{itm}_{x_{i},\hat{y}_{i}}[1]+ \log\mathbf{p}^{itm}_{\hat{x}_{i}, y_{i}}[1]),
\label{eq:loss_itm}
\end{align}
}%
where $\mathbf{p}^{itm}_{x_{i},y_{i}}$ is the two-class probabilities for the $i$-th image-text pair $(x_i, y_i)$. $\hat{y}_{i}$ is a negative text w.r.t. image $x_i$ and $\hat{x}_{i}$ is a negative image w.r.t. text $y_i$. 
The negatives can be randomly sampled~\cite{lu2019vilbert,li2020oscar,kim2021vilt} or mined from hard negatives based on similarity scores~\cite{chen2019uniter,lu2019vilbert,li2021align}. 
In addition, during pre-training, cross encoders are often trained with a Masked Language Model (MLM) loss~\cite{devlin2018bert} $\mathcal{L}_{mlm}$.

\begin{figure*}[!t]
  \centering
  \includegraphics[width=0.96\linewidth]{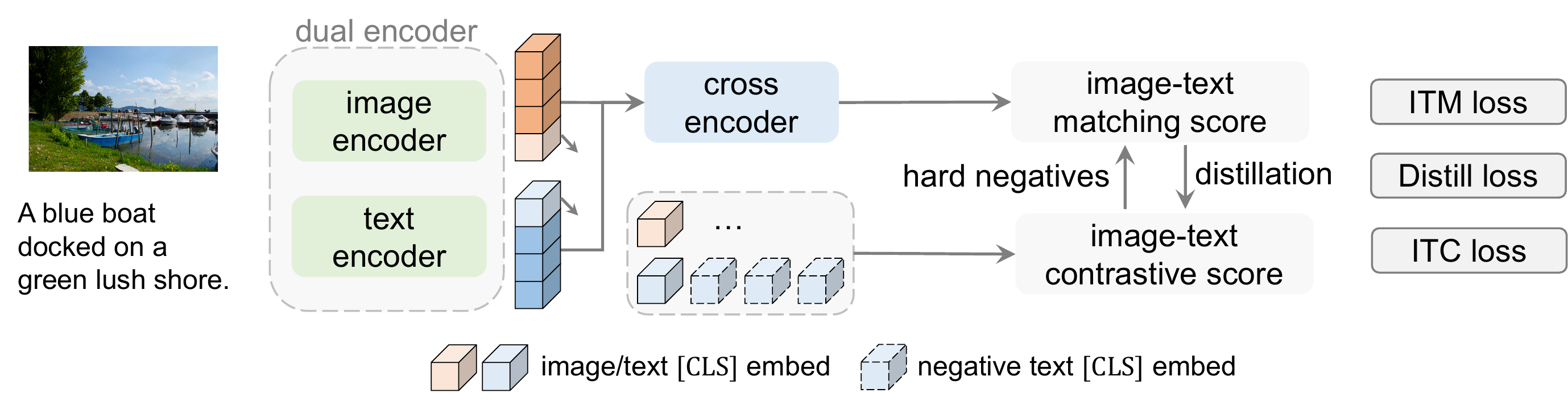}
  \caption{Our implementation of \ModelName. It consists of an \textit{image encoder} and a \textit{text encoder} as its \textit{dual encoder}, and a \textit{cross encoder} on the top. All the encoders are implemented as transformers. The dual encoder encodes input image and text into their respective embedding sequences, and computes Image-Text Contrastive (ITC) scores via dot product of the \texttt{[CLS]} embeddings. The embedding sequences are also forwarded into the cross encoder to compute Image-Text Matching (ITM) scores of positive and in-batch hard negative (from ITC) image-text pairs. As cross encoder is typically more powerful than the dual encoder, we distill the knowledge from the ITM scores to the ITC scores via an online distillation loss. For brevity, MLM loss is not shown in this figure. See details in Section~\ref{sec:method}.
  }
  \vspace{-10pt}
  \label{fig:model_overview}
\end{figure*}

\paragraph{\ModelName: Combined Dual Encoder and Cross Encoder.}
Dual encoders and cross encoders are often seen as separate architectures. 
In \ModelName~(shown in Figure~\ref{fig:model_overview}), we combine them together. 
Specifically, we use a dual encoder to produce aligned representations as the inputs to the cross encoder, i.e., the cross encoder's matching scores are now computed as $h(x,y) = \mathcal{H}(\mathcal{F}(x), \mathcal{G}(y))$. 
It has been shown in~\cite{li2021align} that such aligned representations help improve the cross encoder. 
The negatives in the cross encoder's ITM objective $\mathcal{L}^{itm}$ in Equation~\ref{eq:loss_itm} are mined online~\cite{li2021align} based on the dual encoder's image-text contrastive scores. 

Due to deeper interaction, the cross encoder contains more accurate matching signals than the dual encoder. 
We propose to distill the knowledge from the cross encoder teacher to the dual encoder student. 
The standard distillation objective~\cite{hinton2015distilling} is defined for classification tasks, where a teacher probability distribution $\mathbf{q} \in \mathbb{R}^c$ over the $c$ classes are used as soft labels for learning student probability distribution $\mathbf{p} \in \mathbb{R}^c$, via cross entropy $\mathrm{H}(\mathbf{p}, \mathbf{q})$.
However, such distribution is not directly available for image-text retrieval.

To address this issue, we construct an image negative set and a text negative set.
Specifically, given a matched image-text pair $(x_i, y_i)$, we take $m$ in-batch text hard negatives and pair it with the image $\{(x_i, \hat{y_i})\}$, and denote $\mathcal{N}^{txt}_{i} {=} \{(x_i, y_i)\} \cup \{(x_i, \hat{y_i})\}$ with $|\mathcal{N}^{txt}_{i}|{=}m+1$ as the set of image-text pairs with the text hard negatives.
Thus we can define the predicted probability distribution $\mathbf{p}^{txt}_{i} \in \mathbb{R}^{m+1}$ over $\mathcal{N}^{txt}_{i}$ from the dual encoder, with $p^{txt}_{i,k}$ being its $k$-th element, denoting the probability of the $k$-th pair in $\mathcal{N}^{txt}_{i}$ being the matched pair. $p^{txt}_{i,k}$ is computed as:
\begin{align}
    p^{txt}_{i,k} = \frac{\exp(s_{x_k, y_k} / \tau)}{\sum_{ (x, y) \in \mathcal{N}^{txt}_{i}}\exp(s_{x, y} / \tau)}.
\end{align}
Note that $s_{x, y}$ is the matching score for image-text pair $(x,y)$ from the dual encoder, and $\tau$ is the same temperature parameter used in Equation~\ref{eq:itc1}.
For the cross encoder teacher, we replace $s(x,y)$ with $h^{pos}_{x,y}$:
\begin{align}
    q^{txt}_{i,k} = \frac{\exp(h^{pos}_{x_k, y_k} / \tau)}{\sum_{ (x, y) \in \mathcal{N}^{txt}_{i}}\exp(h^{pos}_{x, y} / \tau)},
\end{align}
where $q^{txt}_{i,k}$ is the $k$-th element of the cross encoder score distribution  $\mathbf{q}^{txt}_{i} \in \mathbb{R}^{m+1}$. 
The loss is calculated as:
\begin{align}
    \mathcal{L}_{distill}^{txt} = \sum_{i=1}^{n} \mathrm{H}(\mathbf{p}^{txt}_{i}, \mathrm{stop}{\text -}\mathrm{grad}(\mathbf{q}^{txt}_{i})).
    \label{eq:distill}
\end{align}
Since our teacher and student network are trained together within the same model, we use $\mathrm{stop}{\text -}\mathrm{grad}$ to indicate stopping gradients from back-propagating to $\mathbf{q}^{txt}_{i}$, and thus the cross encoder.
This is crucial as it prevents the outputs of the network (i.e., the inputs to the loss) from ``collapsing''.
A similar idea has been exploited in the siamese network for  representation learning~\cite{chen2021exploring}.

The final distillation loss is a simple summation of these two terms $\mathcal{L}_{distill}=\mathcal{L}_{distill}^{txt}+\mathcal{L}_{distill}^{img}$. 
Besides, we also use Masked Language Modeling~\cite{devlin2018bert} loss during pretraining.
Overall, the model is trained with the sum of losses:
\begin{align}
    \mathcal{L} =\mathcal{L}_{itc} + \mathcal{L}_{itm} + \mathcal{L}_{mlm} + \mathcal{L}_{distill}.
    \label{eq:final_loss}
\end{align}

Prior work~\cite{miech2021thinking} also proposed a distillation objective for cross encoder to dual encoder knowledge transfer. Our formulation differs from it in the following aspects: 
($i$) \textit{negative types}: we construct the negative image-text pairs for both image and text, instead of only constructing for text. In Section~\ref{subsec:distill_ablation} we empirically demonstrate that our symmetric formulation gives the best overall performance. 
($ii$) \textit{negative sampling method}: \cite{miech2021thinking}~samples all the other examples within the batch as negatives for distillation, while we use only the top-$m$ hard negatives ranked by the dual encoder. In Section~\ref{subsec:distill_ablation} we show that the performance saturates at as few as 4 hard negatives, indicating that only the hard negatives are informative in the distillation process. 
This has an important implication -- that we need only compute the cross encoder scores for the top-$m$ hard negatives, making the computation more efficient.
In the extreme case of using only a single hard negative, we can directly take the inputs to $\mathcal{L}_{itm}$ as teacher scores \textit{without extra cross encoder computation}. 
($iii$) \textit{online vs. offline teacher}: we use an online teacher that is trained together with the student in the same model, while~\cite{miech2021thinking} follows the standard distillation~\cite{hinton2015distilling} to use a separate offline teacher. 
In Section~\ref{subsec:distill_ablation} we show that distillation from an online teacher works well. Moreover, our training can be performed in a single run without an extra run for learning the teacher model.

\paragraph{Model Architecture.}
We show an overview of our implementation of the proposed \ModelName~architecture in Figure~\ref{fig:model_overview}. 
Our dual encoder consists of a separate image encoder and a separate text encoder. 
The image encoder is a 12-layer ViT model~\cite{dosovitskiy2020image}, initialized using weights from a DeiT~\cite{touvron2021training} model trained on ImageNet~\cite{deng2009imagenet} 1K dataset.
The text encoder is a 6-layer transformer encoder~\cite{vaswani2017attention} initialized from the first 6 layers' weights of BERT-base~\cite{devlin2018bert}.
Each transformer layer in the image and text encoders mainly consists of a self-attention layer and a feedforward net (FFN).
After the input image and text are encoded, we take their \texttt{[CLS]} token presentations to compute dual encoder's ITC scores.

The cross encoder is a 6-layer transformer encoder, each layer mainly consisting of a self-attention layer, a cross-attention layer and an FFN. 
It is initialized from the last 6-layer of BERT-base.
The text output from the text encoder is forwarded as input to the cross encoder with the image output injected via the cross-attention layer as the key and value.
The ITM score is computed based on the \texttt{[CLS]} token representation from the last layer.

\section{Experiments} \label{sec:experiments}

\subsection{Dataset and Implementation Details}\label{subsec:implementation_details}
\paragraph{Pre-Training.} 
We pre-train our model on four popular image-text datasets, COCO Captions~\cite{lin2014microsoft,chen2015microsoft}, Visual Genome (VG) Captions~\cite{krishna2017visual}, SBU Captions~\cite{ordonez2011im2text} and Conceptual Captions (CC)~\cite{sharma2018conceptual}, with in total around 4M images. 
This is the same set of datasets used in~\cite{chen2019uniter,sun2021lightningdot,li2021align} for pre-training.
Our model is pre-trained for 30 epochs with a learning rate of 1e-4, weight decay 0.02. 
We use AdamW~\cite{loshchilov2017decoupled} optimizer and warm up the learning rate in the first 1,000 iterations followed by cosine decay to 1e-5.
We also warm up the weight of the distillation loss in the first epoch from 0 to 1, and keep it to be 1 for the rest of the training. 
The number of negatives $m$ is set to 4, the initial value of trainable temperature parameter $\tau$ is set to 0.07.
We use input image resolution of 256\x256 and apply RandAugment~\cite{cubuk2020randaugment} (without color augmentations). The batch size is set to 64 per GPU, and we use 8 NVIDIA A100 GPUs for training. 
The training takes around 2.5 days.

\paragraph{Downstream.}
After pre-training, we fine-tune our model on COCO~\cite{lin2014microsoft,chen2015microsoft} and Flickr30K~\cite{young2014image} for image-text retrieval.
For COCO, we use the Karpathy split~\cite{karpathy2015deep}, which contains 113K training images, 5k validation images, and 5k test images. 
For Flickr30K, we use the standard split with 29K training images, 1K validation images, and 1K test images.
Each image in these two datasets is paired with 5 captions.
During fine-tuning, for both datasets, we use a batch size of 512 with input image resolution 512\x512 as in~\cite{chen2021learning}, and a peak learning rate of 1e-5.
We train the model for 10 epochs for COCO and 25 epochs for Flickr30K.

\subsection{Architecture and Distillation Ablations}\label{subsec:distill_ablation}

In this section, we study the design of this unified architecture, putting our focus on the distillation objective. 
If not otherwise stated, we fine-tune the models on COCO karpathy train split~\cite{karpathy2015deep} from the same pre-trained checkpoint trained on COCO+VG+SBU+CC without distillation, and report R@1 on the 5K val split.
The default input image resolution is set to 384\x384. We use a batch size of 256 and train the models for 5 epochs.

\paragraph{How many negatives should we use?}
During distillation, for each text caption, we construct a bag of $m$ hard negative images, together with its matched image to form a distribution. 
Similarly, we construct another distribution with $m$ hard negative text captions and a positive text caption for each image. 
To understand how the number of negatives $m$ affects the model performance, we evaluate model variants that use $m \in \{1, 4, 9, 14, 31\}$ negative examples.
The results are shown in Table~\ref{tab:distill_num_neg}.
Compared to the base model that is fine-tuned without distillation, we notice the performance of models with distillation improve significantly for dual encoder in both text retrieval and image retrieval, while for cross encoder, the results stay similar.
Note that in this experiment, distillation is only applied during fine-tuning (to save computation), we expect more significant improvement when distillation is used in both pre-training and fine-tuning, as we discuss later in this section.

It is also worth noting that, while prior work~\cite{miech2021thinking} uses all in-batch examples ($m$=31) as negatives, here we show that using a single hard negative example ($m$=1) already provides a notable performance gain for dual encoder (without introducing extra cross encoder forward cost as it reuses the negative score computed for $\mathcal{L}_{itm}$), and the model performance saturates when we use 4 negatives.\footnote{Using $m$=4 is 1.4\x~faster in training than using all negatives ($m$=31).}
Since cross encoder computation is quite expensive, this makes our approach more efficient than~\cite{miech2021thinking}.

\begin{table}[t]
\centering
\small
\tablestyle{5pt}{1.2}
\begin{tabular}{c|cc|cc|c}
& \multicolumn{2}{c|}{cross encoder} & \multicolumn{2}{c|}{dual encoder} & \#extra fwd  \\
\#negatives $m$ & TR & IR & TR & IR & pairs \\
\shline
- (w/o distillation) & 74.98 & 57.41 & 62.64 & 46.78 & 0 \\
1 & \bf 75.02 & 57.44 & 63.10 & 47.67 & 0 \\
4 & 75.00 & \bf 57.76 & 63.64 & \bf 47.99 & 3 \\
9 & 74.92 & 57.63 & \bf 63.82 & 47.74 & 8 \\
14 & 74.96 & 57.63 & 63.20 & 47.95 & 13 \\
31 (all negatives) & 74.70 & 57.31 & 63.60 & 47.92 & 30 \\
\end{tabular}

\caption{\textbf{Effect of number of negatives in distillation} (R@1 on COCO 5k val split). \textit{TR}=Text Retrieval, \textit{IR}=Image Retrieval. 
\textit{\#extra fwd pairs} denotes the number of extra negative image-text pairs that are used in the cross encoder's forward pass for computing distillation teacher score.
As we can reuse one negative pair score computed for $\mathcal{L}_{itm}$, \textit{\#extra fwd pairs} equals to $m$-1. 
}
\vspace{-5pt}
\label{tab:distill_num_neg}
\end{table}

\begin{table}[t]
\centering
\small
\tablestyle{8pt}{1.2}
\begin{tabular}{c|cc|cc}
\multicolumn{1}{c|}{negative sampling} & \multicolumn{2}{c|}{cross encoder} & \multicolumn{2}{c}{dual encoder} \\
\multicolumn{1}{c|}{method} & TR & IR & TR & IR \\
\shline
- & 74.98 & 57.41 & 62.64 & 46.78 \\
random & 74.96 & 57.55 & 62.86 & 46.83 \\
hard & \bf 75.00 & \bf 57.76 & \bf 63.64 & \bf 47.99 \\
\end{tabular}

\caption{\textbf{Effect of negative sampling method in distillation.}
We use 4 negatives (i.e., $m$=4) for all the models with distillation.
}
\label{tab:distill_neg_sampling}
\vspace{-10pt}
\end{table}

\paragraph{Hard negatives are more informative.} 
Our distillation strategy only considers top-$m$ negatives to construct the distribution. 
In Table~\ref{tab:distill_neg_sampling} we compare with an alternative approach that uses $m$ random negatives for distillation. 
We notice that while using random negatives gives slightly better performance than not using distillation, it performs significantly worse than the model that uses hard negatives. 
This makes sense in that the dual encoder is already powerful enough to clearly distinguish positive examples w.r.t. most of the negative examples, and it only struggles with a few hard negatives.
Therefore hard negatives are more informative than random negatives in the distillation process.

\paragraph{Both image and text negatives are important for distillation.}
Our distillation involves distillation from two directions, one from the image's perspective and another from the text's perspective.
For example, for each image, we construct a set of negative text captions along with its matched positive text caption to form a distribution for distillation, the corresponding loss is written as $\mathcal{L}^{txt}_{distill}$ in Section~\ref{sec:method}. 
Similarly, we have $\mathcal{L}^{img}_{distill}$ for each text with its negative images. In~\cite{miech2021thinking}, only image negatives are considered for distillation. 
Here, we study how these two types of negatives affect the model performance. The results are shown in Table~\ref{tab:distill_neg_type}.
From the table, we notice that adding image negatives significantly improves the image retrieval performance of the dual encoder, while adding text negatives boosts the performance of text retrieval. 
The best overall performance is achieved by combining both image and text negatives.

\begin{table}[t]
\centering
\small
\tablestyle{8pt}{1.2}
\begin{tabular}{l|cc|cc}
& \multicolumn{2}{c|}{cross encoder} & \multicolumn{2}{c}{dual encoder} \\
negative type & TR & IR & TR & IR \\
\shline
-  & 74.98 & 57.41 & 62.64 & 46.78 \\
image ($\mathcal{L}^{img}_{distill}$) & \bf 75.04 & \bf 57.76 & 61.94 & 47.84 \\
text ($\mathcal{L}^{txt}_{distill}$) & 74.68 & 57.52 & \bf 63.86 & 47.26 \\
image + text & 75.00 & \bf 57.76 & 63.64 & \bf 47.99 \\
\end{tabular}
\caption{\textbf{Effect of negative types in distillation.}
}
\vspace{-2pt}
\label{tab:distill_neg_type}
\end{table}

\begin{table}[t]
\centering
\small
\tablestyle{6pt}{1.2}
\begin{tabular}{cc|cc|cc}
\multicolumn{2}{c|}{when to distill} & \multicolumn{2}{c|}{cross encoder} & \multicolumn{2}{c}{dual encoder} \\
pre-training & fine-tuning & TR & IR & TR & IR \\
\shline
- & - & 74.98 & 57.41 & 62.64 & 46.78 \\
- & \cmark & 75.00 & 57.76 & 63.64 & 47.99 \\
\cmark & - & \bf 75.02 & 57.67 & 64.70 & 49.96 \\
\cmark & \cmark & 74.96 & \bf 57.82 & \bf 65.00 & \bf 50.53 \\
\end{tabular}
\caption{\textbf{Effect of using distillation at different stages.} 
}
\label{tab:distill_stage}
\end{table}

\paragraph{Should we use distillation in both pre-training and fine-tuning?} 
Current image-text retrieval models, and vision-and-language models~\cite{chen2019uniter,li2020oscar,li2021align} in general, typically follow a two-stage training paradigm: a pre-training stage that trains the model on a large corpus of image-text data (possibly noisy), followed by a fine-tuning stage that tunes the model on a specific downstream task dataset (e.g., COCO Captions~\cite{chen2015microsoft}).
Here we study the effect of using the proposed distillation objective either only in pre-training or fine-tuning, or both.
The results are shown in Table~\ref{tab:distill_stage}.

When using distillation in even only one of the stages, we notice a notable performance gain in the dual encoder compared to the model that does not use distillation. 
Another observation is that using distillation in pre-training provides more performance boost than using it only in fine-tuning. 
This is not surprising as previous work~\cite{touvron2021training} shows that distillation typically benefits from training with more data and a longer training schedule.
Overall, the best performance is achieved by the model that uses distillation in both pre-training and fine-tuning.

\begin{table}[t]
\centering
\small
\tablestyle{6pt}{1.2}
\begin{tabular}{cl|cc|cc}
& & \multicolumn{2}{c|}{cross encoder} & \multicolumn{2}{c}{dual encoder} \\
& teacher type & TR & IR & TR & IR \\
\shline
\bf (a) & - & 74.98 & 57.41 & 62.64 & 46.78 \\
\bf (b) & offline from \textbf{(a)} & 74.68 & 57.64 & 63.20 & 47.65 \\
\bf (c) & momentum & 74.88 & 57.61 & 62.98 & 47.84 \\
\bf (d) & online (step=10) & 74.90 & 57.66 & 63.02 & 47.80 \\
\bf (e) & online (step=100) & 74.92 & 57.67 & 63.12 & 47.79 \\
\bf (f) & online & \bf 75.00 & \bf 57.76 & \bf 63.64 & \bf 47.99 \\
\end{tabular}
\caption{\textbf{Effect of using online and offline teacher.} 
}
\vspace{-3pt}
\label{tab:distill_teacher}
\end{table}

\begin{table}[t]
\centering
\small
\tablestyle{6pt}{1.2}
\begin{tabular}{c|cc|cc}
& \multicolumn{2}{c|}{cross encoder} & \multicolumn{2}{c}{dual encoder} \\
& TR & IR & TR & IR \\
\shline
w/o stop-grad & 56.5 & 38.19 & 63.12 & 47.7 \\
w/ stop-grad & \bf 75.0 & \bf 57.76 & \bf 63.64 & \bf 47.99 \\
\end{tabular}

\caption{\textbf{Distillation with and without stop-grad. 
}
}
\label{tab:distill_stop_grad}
\end{table}

\begin{table*}[!t]
\tablestyle{5pt}{1.1} 

\begin{tabular}{lc|cccccc|cccccc}
\multirow{3}{*}{ Method } & \multirow{3}{*}{ \tabincell{c}{\#PT \\ images} } & \multicolumn{6}{c|}{COCO (5K test set)} & \multicolumn{6}{c}{Flickr30K (1K test set)} \\
& & \multicolumn{3}{c}{TR} & \multicolumn{3}{c|}{IR} & \multicolumn{3}{c}{TR} & \multicolumn{3}{c}{IR} \\
& & R@1 & R@5 & R@10 & R@1 & R@5 & R@10 & R@1 & R@5 & R@10 & R@1 & R@5 & R@10 \\
\shline
\demph{ALIGN~\cite{jia2021scaling}} & \demph{1.2B} & \demph{77.0} & \demph{93.5} & \demph{96.9} & \demph{59.9} & \demph{83.3} & \demph{89.8} & \demph{95.3} & \demph{99.8} & \demph{100.0} & \demph{84.9} & \demph{97.4} & \demph{98.6} \\
ThinkingFastSlow~\cite{miech2021thinking} & 3.1M & - & - & - & - & - & - & - & - & - & 72.1 & 91.5 & 95.2 \\
UNITER~\cite{chen2019uniter} & 4M & 65.7 & 88.6 & 93.8 & 52.9 & 79.9 & 88.0 & 87.3 & 98.0 & 99.2 & 75.6 & 94.1 & 96.8 \\
VILLA~\cite{gan2020large} & 4M & - & - & - & - & - & - & 87.9 & 97.5 & 98.8 & 76.3 & 94.2 & 96.8 \\
LightningDot~\cite{sun2021lightningdot} + UNITER & 4M & 64.6 & 87.6 & 93.5 & 50.3 & 78.7 & 87.5 & 86.5 & 97.5 & 98.9 & 72.6 & 93.1 & 96.1 \\
ViLT~\cite{kim2021vilt} & 4M & 61.5 & 86.3 & 92.7 & 42.7 & 72.9 & 83.1 & 83.5 & 96.7 & 98.6 & 64.4 & 88.7 & 93.8 \\
OSCAR~\cite{li2020oscar} & 4M & 70.0 & 91.1 & 95.5 & 54.0 & 80.8 & 88.5 & - & - & - & - & - & - \\
RerankSmart~\cite{geigle2021retrieve} + OSCAR & 4M & 70.8 & 91.0 & 95.2 & 54.7 & 81.3 & 88.0 & 89.4 & 97.7 & 99.0 & 76.4 & 93.6 & 96.2 \\
ALBEF~\cite{li2021align} & 4M & 73.1 & 91.4 & 96.0 & 56.8 & 81.5 & 89.2 & 94.3 & \bf 99.4 & 99.8 & 82.8 & \bf 96.7 & \bf 98.4 \\
\midrule
\ModelName~(cross-enc) & 4M & \bf 75.1 & \bf 92.4 & \bf 96.7 & \bf 58.0 & \bf 82.8 & \bf 89.7 & \bf 94.4 & \bf 99.4 & \bf 99.9 & \bf 83.4 & 96.4 & 98.2 \\
\end{tabular}

\caption{
\textbf{Comparison to state-of-the-art image-text retrieval methods} on COCO and Flickr30k datasets. \demph{Gray} indicates models trained with significantly larger amount of data.
Here we show cross encoder performance for \ModelName.
}
\vspace{-10pt}
\label{tab:sota_comparison_ret_all}
\end{table*}

\begin{table}[t]
\centering
\small
\tablestyle{3pt}{1.2}
\begin{tabular}{l|cc|cc}
& \multicolumn{2}{c|}{cross encoder} & \multicolumn{2}{c}{dual encoder} \\
& TR & IR & TR & IR \\
\shline
\textbf{(a)} base & \bf 74.98 & \bf 57.41 & \bf 62.64 & \bf 46.78 \\
\textbf{(b)} w/o cross encoder ($\mathcal{L}_{itm}$) & - & - & 60.24 & 43.56 \\
\textbf{(c)} w/o dual encoder ($\mathcal{L}_{itc}$) & 65.26 & 50.98 & - & - \\
\textbf{(d)} w/o hard negative for cross encoder & 67.82 & 51.63 & 60.70 & 46.09 \\
\end{tabular}
\caption{\textbf{Combining cross encoder and dual encoder.} 
Line \textit{b}-\textit{d} each removes a component from the base model in Line \textit{a}.
}
\label{tab:ablation_cross_dual}
\end{table}

\paragraph{Online \textit{vs.} offline teacher.} 
Standard distillation~\cite{hinton2015distilling} trains a student model with supervision from an offline learnt teacher model which is kept constant during distillation. 
In~\ModelName, we use an online teacher instead, which is dynamically evolving along with the student model. 
We examine the effect of using an online teacher compared to that using a standard offline teacher. 
The results are shown in Table~\ref{tab:distill_teacher}. 
The offline teacher model is the trained model from Table~\ref{tab:distill_teacher}a.
Except for the offline teacher, we also compare with a momentum teacher that updates from the online model every step with a momentum of 0.995 (Table~\ref{tab:distill_teacher}c), and two online model variants that update the teacher model by copying the online model every 10 or 100 steps (Table~\ref{tab:distill_teacher}c,d, they can also be seen as momentum models with momentum 0, and update every 10 or 100 steps).
From the table we see that all distillation variants improve the performance of the baseline model in Table~\ref{tab:distill_teacher}a, showing that distillation is useful in all the cases. 
The model with offline teacher performs similarly to the model with a momentum teacher, while both being worse than the model with online teacher.

\paragraph{The role of stop-grad in Equation~\ref{eq:distill}.}
In visual representation learning, SimSiam~\cite{chen2021exploring} suggests that \texttt{stop-grad} is important for siamese network from collapsing to a trivial solution.
Since predictions and targets in our distillation formulation are also from the same network, it may suffer the same collapsing as siamese networks.
Therefore, in Equation~\ref{eq:distill} we also use \texttt{stop-grad} to disable the gradients flow from the distillation loss to the cross encoder. 
To study the effect of using \texttt{stop-grad}, we compare a variant that does not use \texttt{stop-grad} in Table~\ref{tab:distill_stop_grad}. 
We notice that, though the model performance does not collapse to a random number as in SimSiam, we do observe performance drop for both encoders, and especially for the cross encoder, where the performance drops almost 20 points.
Here we do not see a severe collapse as in SimSiam, probably because our model is also trained with other objectives (e.g., $\mathcal{L}_{itc}$ and $\mathcal{L}_{itm}$) instead of a single one in SimSiam.

\paragraph{The effect of combining dual encoder and cross encoder.}
\ModelName~combines dual and cross encoder in the same architecture, optimized jointly. 
To examine how they affect each other, we compare the base model (without distillation) with the variants that removed the losses associated with one of them. 
The results are shown in Table~\ref{tab:ablation_cross_dual}.
Table~\ref{tab:ablation_cross_dual}b removes the cross encoder and only trains the dual encoder part, its dual encoder performance drops significantly compared to the base model in Table~\ref{tab:ablation_cross_dual}a, suggesting that cross encoder objective is also useful in learning a better dual encoder.
We hypothesize that the dual encoder benefits from the fine-grained alignment learned in the cross encoder.

Similarly, when removing the dual encoder objective $\mathcal{L}_{itc}$ in Table~\ref{tab:ablation_cross_dual}c, we observe a notable performance drop on the cross encoder. 
While cross encoder is able to perform dense and fine-grained interactions and alignment between the two modalities, it does so implicitly, which can be harder to learn.
In contrast, the dual encoder performs explicit alignment between the input image and text, and using better-aligned representation as inputs helps the cross encoder to focus more on fine-grained details.

In addition, we also study the effect of using hard negatives for cross encoder's $\mathcal{L}_{itm}$ objective. We compare our base model (Table~\ref{tab:ablation_cross_dual}a which uses hard negatives provided by the dual encoder) with a variant (Table~\ref{tab:ablation_cross_dual}d) that uses random negatives.
We notice that when switching from hard negatives to random negatives, the performance of both cross and dual encoders drops, and the drop is especially significant for cross encoder.
This suggests that using hard negatives is crucial for cross encoder's performance, and a better cross encoder has a positive impact on the dual encoder.

\paragraph{Summary:}
($i$) Distillation from the cross encoder teacher to the dual encoder student is useful in improving the dual encoder's performance; 
($ii$) The distillation objective is effective using as few as a single negative (i.e., $m$=1); 
($iii$) Using hard negatives as the anchor for distillation is crucial as they are more informative than random negatives; 
($iv$) Image negatives are more important in improving image retrieval performance while text negatives are more useful in improving text retrieval performance. 
Combining both types of negatives gives the best overall performance; 
($v$) Distillation is helpful either used during pre-training or fine-tuning, and it is more helpful if used in both stages; 
($vi$) Applying \texttt{stop-grad} on cross encoder teacher's score is important to prevent the model from collapsing; 
($vii$) Online teacher gives better performance than an offline teacher or a momentum teacher; 
($viii$) Dual encoder and cross encoder help improve each other when trained together, and hard negatives are important for cross encoder learning.

\begin{table*}[!t]
\tablestyle{5pt}{1.1} 

\begin{tabular}{lc|cccccc|cccccc}
\multirow{3}{*}{ Method } & \multirow{3}{*}{ \tabincell{c}{\#PT \\ images} } & \multicolumn{6}{c|}{COCO (5K test set)} & \multicolumn{6}{c}{Flickr30K (1K test set)} \\
& & \multicolumn{3}{c}{TR} & \multicolumn{3}{c|}{IR} & \multicolumn{3}{c}{TR} & \multicolumn{3}{c}{IR} \\
& & R@1 & R@5 & R@10 & R@1 & R@5 & R@10 & R@1 & R@5 & R@10 & R@1 & R@5 & R@10 \\
\shline
\demph{ALIGN~\cite{jia2021scaling}} & \demph{1.2B} & \demph{77.0} & \demph{93.5} & \demph{96.9} & \demph{59.9} & \demph{83.3} & \demph{89.8} & \demph{95.3} & \demph{99.8} & \demph{100.0} & \demph{84.9} & \demph{97.4} & \demph{98.6} \\
\demph{VSE$\infty$\cite{chen2021learning}} & \demph{1B} & \demph{66.4} & \demph{89.3} & - & \demph{51.6} & \demph{79.3} & - & \demph{88.4} & \demph{98.3} & \demph{99.5} & \demph{74.2} & \demph{93.7} & \demph{96.8} \\
LightningDot~\cite{sun2021lightningdot} & 4M & 60.1 & 85.1 & 91.8 & 45.8 & 74.6 & 83.8 & 83.9 & 97.2 & 98.6 & 69.9 & 91.1 & 95.2 \\
RerankSmart~\cite{geigle2021retrieve} $^{*}$ & 4M & 66.9 & 90.1 & 95.0 & \bf 52.2 & \bf 80.2 & \bf 88.0 & 86.3 & 96.8 & 98.6 & 71.6 & 91.5 & 95.0 \\
\midrule
\ModelName~(dual-enc) & 4M & \bf 67.6 & \bf 90.5 & \bf 95.4 & 51.7 & 79.2 & 87.5 & \bf 89.6 & \bf 98.6 & \bf 99.5 & \bf 77.2 & \bf 94.3 & \bf 97.6 \\
\end{tabular}

\caption{
\textbf{Comparison to dual encoder image-text retrieval methods} on COCO and Flickr30k datasets. $^{*}$ the COCO performance for RerankSmart may benefit from pre-training on extra datasets (VQA~\cite{antol2015vqa} and GQA~\cite{hudson2019gqa}) annotated on COCO images. For VSE$\infty$, we specify \#images used in image-only pre-training, while for other models, we specify \#images used in image-text pre-training. 
}
\vspace{-10pt}
\label{tab:sota_comparison_ret_dual_enc}
\end{table*}

\begin{table}[!t]
\tablestyle{2pt}{1.1} 

\begin{tabular}{lc|cccccc}
\multirow{3}{*}{ Method} & \multirow{3}{*}{ \tabincell{c}{\#PT\\images} } & \multicolumn{6}{c}{Flickr30K (1K test set)} \\
& & \multicolumn{3}{c}{TR} & \multicolumn{3}{c}{IR} \\
& & R@1 & R@5 & R@10 & R@1 & R@5 & R@10 \\
\shline
\demph{ALIGN~\cite{jia2021scaling}} & \demph{1.2B} & \demph{88.6} & \demph{98.7} & \demph{99.7} & \demph{75.7} & \demph{93.8} & \demph{96.8} \\
\demph{CLIP~\cite{radford2021learning}} & \demph{400M} & \demph{88.0} & \demph{98.7} & \demph{99.4} & \demph{68.7} & \demph{90.6} & \demph{95.2} \\
UNITER~\cite{chen2019uniter} & 4M & 80.7 & 95.7 & 98.0 & 66.2 & 88.4 & 92.9 \\
OSCAR~\cite{li2020oscar} & 4M & 81.0 & 95.5 & 97.8 & 67.2 & 88.5 & 92.7 \\
RerankSmart~\cite{geigle2021retrieve} & 4M & 78.2 & 94.0 & 97.3 & 63.3 & 86.4 & 91.6 \\
ALBEF~\cite{li2021align} & 4M & 90.5 & 98.8 & 99.7 & 76.8 & 93.7 & 96.7 \\
\midrule
\ModelName~(dual) & 4M & 82.3 & 96.7 & 98.8 & 70.3 & 91.2 & 95.6 \\
\ModelName~(cross) & 4M & \bf 91.8 & \bf 99.0 & \bf 100.0 & \bf 79.2 & \bf 94.4 & \bf 97.1 \\
\end{tabular}

\caption{
\textbf{Zero-shot retrieval results} on Flickr30K.
}
\vspace{-8pt}
\label{tab:sota_comparison_ret_zero_shot}
\end{table}

\subsection{Comparison to State-of-the-art}

In Table~\ref{tab:sota_comparison_ret_all}, we compare \ModelName's cross encoder to state-of-the-art approaches on COCO and Flickr30K for image-text retrieval. 
We notice our model achieves strong performance on both text retrieval (\textit{TR}) and image retrieval (\textit{IR}). 
The improvement over prior state-of-the-art is more significant on COCO compared to that of Flickr30K, possibly due to the performance of the latter is almost saturated.

In Table~\ref{tab:sota_comparison_ret_dual_enc}, we compare \ModelName's dual encoder with the other dual encoder methods.
Our model shows better overall performance on Flickr30K and comparable performance on COCO.
The slightly lower performance on COCO image retrieval compared to RerankSmart~\cite{geigle2021retrieve} might because its pre-training uses extra datasets (VQA~\cite{antol2015vqa} and GQA~\cite{hudson2019gqa}) annotated on COCO images.

In Table~\ref{tab:sota_comparison_ret_zero_shot}, we show zero-shot results on Flickr30K, for both dual and cross encoder (from the same trained \ModelName).
\ModelName's cross encoder achieves the best overall performance, and is even better than ALIGN~\cite{jia2021scaling} and CLIP~\cite{radford2021learning} which are trained on a significantly larger amount of data.
Meanwhile, thanks to distillation, \ModelName's dual encoder is also better than multiple strong cross encoder methods, e.g., OSCAR~\cite{li2020oscar} and UNITER~\cite{chen2019uniter}, and achieves comparable image retrieval performance to CLIP, showing the benefit of leveraging a strong cross encoder to improve dual encoder.
On the other hand, the performance of the dual encoder of our model still lags behind the cross encoder, especially on R@1. We hypothesize this difference mainly comes from the dual encoder's inability of performing fine-grained matching required for accurate retrieval.

\paragraph{VQA results.}
In addition to image-text retrieval, we also validate the model's effectiveness on VQA~\cite{antol2015vqa}. 
To adapt our model for VQA, we add an answer decoder to generate the answers as natural language text~\cite{cho2021unifying,li2021align}.
We use an input image resolution of 512x512 with a batch size of 256, and we fine-tune the model for 8 epochs with a peak learning rate of 2e-5.
In Table~\ref{tab:sota_comparison_vqa}, we show our model demonstrates strong results compared to previous approaches.

\begin{table}[!t]
\tablestyle{8pt}{1.1} 

\begin{tabular}{l|cc}
Method & test-dev & test-std \\
\shline
VL-BART~\cite{cho2021unifying} & - & 71.3 \\
LXMERT~\cite{tan2019lxmert} & 72.42 & 72.54 \\
UNITER~\cite{chen2019uniter} & 72.70 & 72.91 \\
OSCAR~\cite{li2020oscar} & 73.16 & 73.44 \\
VILLA~\cite{gan2020large} & 73.59 & 73.67 \\
UNIMO~\cite{li2020unimo} & 73.79 & 74.02 \\
ViLT~\cite{kim2021vilt} & 70.94 & - \\
ALBEF~\cite{li2021align} & 74.54 & 74.70 \\
\midrule
\ModelName & \bf 75.18 & \bf 75.20 \\
\end{tabular}

\caption{
\textbf{Results on VQA dataset}~\cite{antol2015vqa}. 
}
\vspace{-5pt}
\label{tab:sota_comparison_vqa}
\end{table}

\section{Discussion and Conclusion} \label{sec:conclusion}
Previous approaches often view dual encoders~\cite{faghri2017vse++,radford2021learning,chen2021learning} and cross encoders~\cite{chen2019uniter,li2020oscar,li2021align} as separate architectures for modeling. 
In this work, we propose a unified approach that combines the two types of architectures in the same network for image-text retrieval. 
We conduct a comprehensive ablation study around this unified architecture and especially lay our focus on the design of the distillation objective -- which is employed to efficiently transfer the knowledge from cross encoder to dual encoder in an online manner.
We empirically demonstrate the effectiveness of joint training of these two architectures, and show that distillation is quite effective in improving dual encoder's performance.
We hope our work will encourage future explorations on combining dual and cross encoders for image-text retrieval, and vision-and-language learning in general.

\paragraph{Limitations:} 
While our distillation objective is more efficient than~\cite{miech2021thinking}, it still induces an additional 6\% computation cost at training with $m$=4 (it does not hurt inference speed). 
This cost is neglectable when we set $m$=1, but it underperforms the setting of $m$=4. 
One interesting future work could be exploring the extreme case of $m$=1.

\paragraph{Societal Impact:} 
The predictions from the developed system reflect the distribution of the data used for training, and they can be inaccurate and biased by the data. 
Therefore, users should not completely rely on our system for making real-world decisions. 

\paragraph{Acknowledgement.}  This research was partially done when Jie was an intern with Meta AI, and was later supported at UNC by ARO Award W911NF2110220 and DARPA MCS Grant N66001-19-2-4031. The views contained in this article are those of the authors and not of the funding agency.

\appendix
\section{Additional Experiments}
\paragraph{Comparison on COCO 5K test set using CxC annotations.} 
In Table~\ref{tab:cxc} we compare \ModelName~with baseline methods on COCO 5k test split with CxC annotations~\cite{parekh2020crisscrossed}. 
CxC augments the original COCO dataset with human semantic similarity judgments for 267,095 intra- and inter-modality pairs. 
We note from the table that our method achieves competitive performance even compared to strong method like VSE$\infty$ that leverages visual encoder trained on 1B images.

\paragraph{Query time comparison.}
Comparing to the efficient cross encoder ViLT and strong dual encoder VSE$\infty$ in Table~\ref{tab:efficiency}, \ModelName~(dual) gives better results while runs faster. Our full cross encoder (Table~\ref{tab:efficiency}e) greatly improves the dual encoder results, though with increased time cost (still faster than ViLT). With reranking (Table~\ref{tab:efficiency}d),\footnote{In this setting, we use cross encoder to rerank the top k retrieved candidates from the dual encoder, as in~\cite{li2021align}. As reranking speeds up inference without hurting performance, we report `cross encoder' results in this setting if not otherwise stated.} we notice a large speedup without performance degradation.

\begin{table}[t]
\centering
\small
\tablestyle{5pt}{1.1}
\begin{tabular}{l|cccc}
Method & I $\rightarrow$ T & T $\rightarrow$ I & T $\rightarrow$ T & I $\rightarrow$ I \\
\shline
\demph{ALIGN (1.2B images)~\cite{jia2021scaling}} & \demph{78.1} & \demph{61.8} & \demph{45.4} & \demph{49.4} \\
\demph{VSE$\infty$ (1B images)~\cite{chen2021learning}} & \demph{67.9} & \demph{53.6} & \demph{46.7} & \demph{51.3} \\
DE~\cite{parekh2020crisscrossed} & 55.9 & 41.7 & 42.6 & 38.5 \\
\midrule
\ModelName~(dual-encoder) & 69.2 & 53.5 & 44.7 & 46.7 \\
\end{tabular}
\caption{R@1 results on COCO 5k test, with CxC~\cite{parekh2020crisscrossed}.
}
\label{tab:cxc}
\end{table}

\begin{table}[t]
\centering
\small
\tablestyle{3pt}{1.1}
\begin{tabular}{l|c|cc|c}
Method & encode type & TR & IR & time \\
\shline
\textbf{(a)} ViLT~\cite{kim2021vilt} & cross & 83.5 & 64.4 & 6613s \\
\textbf{(b)} \demph{VSE$\infty$ (1B images)~\cite{chen2021learning}} & \demph{dual} & \demph{88.4} & \demph{74.2} & \demph{45s} \\
\textbf{(c)} \ModelName~(dual) & dual & 89.6 & 77.2 & \bf 11s \\
\textbf{(d)} \ModelName~(cross) rerank top 16 & dual+cross & \bf 94.5 & \bf 83.4 & 76s \\
\textbf{(e)} \ModelName~(cross) no rerank & cross & 94.4 & 83.1 & 2342s \\
\end{tabular}
\caption{Query time comparison on the full Flickr 1K test split. Time cost is calculated with a single A100 GPU.}
\label{tab:efficiency}
\end{table}

\begin{table}[t]
\centering
\small
\tablestyle{8pt}{1.2}
\begin{tabular}{l|l}
config & value \\
\shline
optimizer & AdamW \\
base learning rate & 1e-4 \\
min learning rate & 1e-5 \\
weight decay & 0.02 \\
optimizer momentum & $\beta_{1},\beta_{2}$=0.9,0.999 \\
batch size & 512 \\
learning rate schedule & cosine decay~\cite{loshchilov2016sgdr} \\
warmup iterations & 2,000 \\
image augmentation & RandAug(N=2, M=7)~\cite{cubuk2020randaugment} \\
training epochs & 30 \\
\end{tabular}
\caption{\textbf{Pre-Training settings.}
}
\label{tab:pretrain_settings}
\end{table}

\begin{table}[t]
\centering
\small
\tablestyle{8pt}{1.2}
\begin{tabular}{l|l}
config & value \\
\shline
optimizer & AdamW \\
base learning rate & 1e-5 \\
min learning rate & 1e-6 \\
weight decay & 0.02 \\
optimizer momentum & $\beta_{1},\beta_{2}$=0.9,0.999 \\
batch size & 512 \\
learning rate schedule & cosine decay~\cite{loshchilov2016sgdr} \\
image augmentation & RandAug (N=2, M=7)~\cite{cubuk2020randaugment} \\
training epochs & 10 \\
\end{tabular}
\caption{\textbf{COCO retrieval fine-tuning settings.}
}
\label{tab:coco_ret_finetune}
\end{table}

\begin{table}[t]
\centering
\small
\tablestyle{8pt}{1.2}
\begin{tabular}{l|l}
config & value \\
\shline
optimizer & AdamW \\
base learning rate & 1e-5 \\
min learning rate & 1e-6 \\
weight decay & 0.02 \\
optimizer momentum & $\beta_{1},\beta_{2}$=0.9,0.999 \\
batch size & 512 \\
learning rate schedule & cosine decay~\cite{loshchilov2016sgdr} \\
image augmentation & RandAug (N=2, M=7)~\cite{cubuk2020randaugment} \\
training epochs & 25 \\
\end{tabular}
\caption{\textbf{Flickr30K retrieval fine-tuning settings.}
}
\label{tab:flickr_ret_finetune}
\end{table}

\begin{table}[t]
\centering
\small
\tablestyle{8pt}{1.2}
\begin{tabular}{l|l}
config & value \\
\shline
optimizer & AdamW \\
base learning rate & 2e-5 \\
min learning rate & 1e-6 \\
weight decay & 0.02 \\
optimizer momentum & $\beta_{1},\beta_{2}$=0.9,0.999 \\
batch size & 512 \\
learning rate schedule & cosine decay~\cite{loshchilov2016sgdr} \\
image augmentation & RandAug (N=2, M=7)~\cite{cubuk2020randaugment} \\
training epochs & 8 \\
\end{tabular}
\caption{\textbf{VQA fine-tuning settings.}
}
\label{tab:vqa_finetune}
\end{table}

\section{Additional Implementation Details}
\paragraph{Pre-Training.} For pre-training, we train the model with distillation $m$=4, and we use both image and text negatives. The image resolution is set to 256\x256. The training follows the default setting in Table~\ref{tab:pretrain_settings}.

\paragraph{Fine-Tuneing.} After pre-training, we fine-tune the model on image-text retrieval task on COCO and Flickr30K, and visual question answering on VQA~\cite{antol2015vqa}.  The image resolution is set to 512\x512 as in~\cite{chen2021learning}, the positional embeddings are interpolated following~\cite{touvron2021training}.
For COCO and Flickr30K retrieval, we train the model with distillation $m$=4, and we use both image and text negatives, and we follow the settings in Table~\ref{tab:coco_ret_finetune} for COCO, Table~\ref{tab:flickr_ret_finetune} for Flickr30K.
For VQA, we follow the settings in Table~\ref{tab:vqa_finetune}.

{\small
\bibliographystyle{ieee_fullname}
\bibliography{egbib}

\begin{thebibliography}{10}\itemsep=-1pt

\bibitem{anne2017localizing}
Lisa Anne~Hendricks, Oliver Wang, Eli Shechtman, Josef Sivic, Trevor Darrell,
  and Bryan Russell.
\newblock Localizing moments in video with natural language.
\newblock In {\em ICCV}, 2017.

\bibitem{antol2015vqa}
Stanislaw Antol, Aishwarya Agrawal, Jiasen Lu, Margaret Mitchell, Dhruv Batra,
  C Lawrence~Zitnick, and Devi Parikh.
\newblock Vqa: Visual question answering.
\newblock In {\em ICCV}, 2015.

\bibitem{ba2013deep}
Lei~Jimmy Ba and Rich Caruana.
\newblock Do deep nets really need to be deep?
\newblock In {\em NeurIPS}, 2013.

\bibitem{bucilua2006model}
Cristian Bucilua, Rich Caruana, and Alexandru Niculescu-Mizil.
\newblock Model compression.
\newblock In {\em KDD}, 2006.

\bibitem{chen2021learning}
Jiacheng Chen, Hexiang Hu, Hao Wu, Yuning Jiang, and Changhu Wang.
\newblock Learning the best pooling strategy for visual semantic embedding.
\newblock In {\em CVPR}, 2021.

\bibitem{chen2020simple}
Ting Chen, Simon Kornblith, Mohammad Norouzi, and Geoffrey Hinton.
\newblock A simple framework for contrastive learning of visual
  representations.
\newblock In {\em ICML}, 2020.

\bibitem{chen2015microsoft}
Xinlei Chen, Hao Fang, Tsung-Yi Lin, Ramakrishna Vedantam, Saurabh Gupta, Piotr
  Doll{\'a}r, and C~Lawrence Zitnick.
\newblock Microsoft coco captions: Data collection and evaluation server.
\newblock {\em arXiv}, 2015.

\bibitem{chen2021exploring}
Xinlei Chen and Kaiming He.
\newblock Exploring simple siamese representation learning.
\newblock In {\em CVPR}, 2021.

\bibitem{chen2021empirical}
Xinlei Chen, Saining Xie, and Kaiming He.
\newblock An empirical study of training self-supervised vision transformers.
\newblock In {\em ICCV}, 2021.

\bibitem{chen2019uniter}
Yen-Chun Chen, Linjie Li, Licheng Yu, Ahmed~El Kholy, Faisal Ahmed, Zhe Gan, Yu
  Cheng, and Jingjing Liu.
\newblock Uniter: Learning universal image-text representations.
\newblock In {\em ECCV}, 2020.

\bibitem{cho2021unifying}
Jaemin Cho, Jie Lei, Hao Tan, and Mohit Bansal.
\newblock Unifying vision-and-language tasks via text generation.
\newblock In {\em ICML}, 2021.

\bibitem{cubuk2020randaugment}
Ekin~D Cubuk, Barret Zoph, Jonathon Shlens, and Quoc~V Le.
\newblock Randaugment: Practical automated data augmentation with a reduced
  search space.
\newblock In {\em CVPR Workshops}, 2020.

\bibitem{deng2009imagenet}
Jia Deng, Wei Dong, Richard Socher, Li-Jia Li, Kai Li, and Li Fei-Fei.
\newblock Imagenet: A large-scale hierarchical image database.
\newblock In {\em CVPR}, 2009.

\bibitem{devlin2018bert}
Jacob Devlin, Ming-Wei Chang, Kenton Lee, and Kristina Toutanova.
\newblock Bert: Pre-training of deep bidirectional transformers for language
  understanding.
\newblock In {\em NAACL}, 2019.

\bibitem{dosovitskiy2020image}
Alexey Dosovitskiy, Lucas Beyer, Alexander Kolesnikov, Dirk Weissenborn,
  Xiaohua Zhai, Thomas Unterthiner, Mostafa Dehghani, Matthias Minderer, Georg
  Heigold, Sylvain Gelly, et~al.
\newblock An image is worth 16x16 words: Transformers for image recognition at
  scale.
\newblock In {\em ICLR}, 2020.

\bibitem{parekh2020crisscrossed}
Parekh et al.
\newblock Crisscrossed captions: Extended intra intermodal semantic similarity
  judgments for ms-coco.
\newblock In {\em EACL}, 2021.

\bibitem{faghri2017vse++}
Fartash Faghri, David~J Fleet, Jamie~Ryan Kiros, and Sanja Fidler.
\newblock Vse++: Improving visual-semantic embeddings with hard negatives.
\newblock In {\em BMVC}, 2018.

\bibitem{fang2021compressing}
Zhiyuan Fang, Jianfeng Wang, Xiaowei Hu, Lijuan Wang, Yezhou Yang, and Zicheng
  Liu.
\newblock Compressing visual-linguistic model via knowledge distillation.
\newblock {\em ICCV}, 2021.

\bibitem{fang2021seed}
Zhiyuan Fang, Jianfeng Wang, Lijuan Wang, Lei Zhang, Yezhou Yang, and Zicheng
  Liu.
\newblock Seed: Self-supervised distillation for visual representation.
\newblock In {\em ICCV}, 2021.

\bibitem{frome2013devise}
Andrea Frome, Greg Corrado, Jonathon Shlens, Samy Bengio, Jeffrey Dean,
  Marc’Aurelio Ranzato, and Tomas Mikolov.
\newblock Devise: A deep visual-semantic embedding model.
\newblock In {\em NeurIPS}, 2013.

\bibitem{gan2020large}
Zhe Gan, Yen-Chun Chen, Linjie Li, Chen Zhu, Yu Cheng, and Jingjing Liu.
\newblock Large-scale adversarial training for vision-and-language
  representation learning.
\newblock In {\em NeurIPS}, 2020.

\bibitem{geigle2021retrieve}
Gregor Geigle, Jonas Pfeiffer, Nils Reimers, Ivan Vuli{\'c}, and Iryna
  Gurevych.
\newblock Retrieve fast, rerank smart: Cooperative and joint approaches for
  improved cross-modal retrieval.
\newblock {\em arXiv}, 2021.

\bibitem{gur-etal-2021-cross-modal}
Shir Gur, Natalia Neverova, Chris Stauffer, Ser-Nam Lim, Douwe Kiela, and
  Austin Reiter.
\newblock Cross-modal retrieval augmentation for multi-modal classification.
\newblock In {\em Findings of EMNLP}, 2021.

\bibitem{he2020momentum}
Kaiming He, Haoqi Fan, Yuxin Wu, Saining Xie, and Ross Girshick.
\newblock Momentum contrast for unsupervised visual representation learning.
\newblock In {\em CVPR}, 2020.

\bibitem{hinton2015distilling}
Geoffrey Hinton, Oriol Vinyals, and Jeff Dean.
\newblock Distilling the knowledge in a neural network.
\newblock In {\em NeurIPS Deep Learning Workshop}, 2014.

\bibitem{howard2017mobilenets}
Andrew~G Howard, Menglong Zhu, Bo Chen, Dmitry Kalenichenko, Weijun Wang,
  Tobias Weyand, Marco Andreetto, and Hartwig Adam.
\newblock Mobilenets: Efficient convolutional neural networks for mobile vision
  applications.
\newblock {\em arXiv}, 2017.

\bibitem{hudson2019gqa}
Drew~A Hudson and Christopher~D Manning.
\newblock Gqa: A new dataset for real-world visual reasoning and compositional
  question answering.
\newblock In {\em CVPR}, 2019.

\bibitem{izacard2020distilling}
Gautier Izacard and Edouard Grave.
\newblock Distilling knowledge from reader to retriever for question answering.
\newblock In {\em ICLR}, 2021.

\bibitem{jia2021scaling}
Chao Jia, Yinfei Yang, Ye Xia, Yi-Ting Chen, Zarana Parekh, Hieu Pham, Quoc~V
  Le, Yunhsuan Sung, Zhen Li, and Tom Duerig.
\newblock Scaling up visual and vision-language representation learning with
  noisy text supervision.
\newblock {\em arXiv}, 2021.

\bibitem{jiao2019tinybert}
Xiaoqi Jiao, Yichun Yin, Lifeng Shang, Xin Jiang, Xiao Chen, Linlin Li, Fang
  Wang, and Qun Liu.
\newblock Tinybert: Distilling bert for natural language understanding.
\newblock In {\em EMNLP}, 2020.

\bibitem{karpathy2015deep}
Andrej Karpathy and Li Fei-Fei.
\newblock Deep visual-semantic alignments for generating image descriptions.
\newblock In {\em CVPR}, 2015.

\bibitem{karpukhin2020dense}
Vladimir Karpukhin, Barlas O{\u{g}}uz, Sewon Min, Patrick Lewis, Ledell Wu,
  Sergey Edunov, Danqi Chen, and Wen-tau Yih.
\newblock Dense passage retrieval for open-domain question answering.
\newblock {\em arXiv}, 2020.

\bibitem{kim2021vilt}
Wonjae Kim, Bokyung Son, and Ildoo Kim.
\newblock Vilt: Vision-and-language transformer without convolution or region
  supervision.
\newblock In {\em ICML}, 2021.

\bibitem{kim2016sequence}
Yoon Kim and Alexander~M Rush.
\newblock Sequence-level knowledge distillation.
\newblock In {\em EMNLP}, 2016.

\bibitem{kiros2014unifying}
Ryan Kiros, Ruslan Salakhutdinov, and Richard~S Zemel.
\newblock Unifying visual-semantic embeddings with multimodal neural language
  models.
\newblock {\em arXiv}, 2014.

\bibitem{krishna2017visual}
Ranjay Krishna, Yuke Zhu, Oliver Groth, Justin Johnson, Kenji Hata, Joshua
  Kravitz, Stephanie Chen, Yannis Kalantidis, Li-Jia Li, David~A Shamma, et~al.
\newblock Visual genome: Connecting language and vision using crowdsourced
  dense image annotations.
\newblock {\em IJCV}, 2017.

\bibitem{lei2021less}
Jie Lei, Linjie Li, Luowei Zhou, Zhe Gan, Tamara~L Berg, Mohit Bansal, and
  Jingjing Liu.
\newblock Less is more: Clipbert for video-and-language learning via sparse
  sampling.
\newblock In {\em CVPR}, 2021.

\bibitem{lei2020tvr}
Jie Lei, Licheng Yu, Tamara~L Berg, and Mohit Bansal.
\newblock Tvr: A large-scale dataset for video-subtitle moment retrieval.
\newblock In {\em ECCV}, 2020.

\bibitem{li2020unicoder}
Gen Li, Nan Duan, Yuejian Fang, Ming Gong, Daxin Jiang, and Ming Zhou.
\newblock Unicoder-vl: A universal encoder for vision and language by
  cross-modal pre-training.
\newblock In {\em AAAI}, 2020.

\bibitem{li2021align}
Junnan Li, Ramprasaath~R Selvaraju, Akhilesh~Deepak Gotmare, Shafiq Joty,
  Caiming Xiong, and Steven Hoi.
\newblock Align before fuse: Vision and language representation learning with
  momentum distillation.
\newblock In {\em NeurIPS}, 2021.

\bibitem{li2020hero}
Linjie Li, Yen-Chun Chen, Yu Cheng, Zhe Gan, Licheng Yu, and Jingjing Liu.
\newblock Hero: Hierarchical encoder for video+ language omni-representation
  pre-training.
\newblock In {\em EMNLP}, 2020.

\bibitem{li2019visualbert}
Liunian~Harold Li, Mark Yatskar, Da Yin, Cho-Jui Hsieh, and Kai-Wei Chang.
\newblock Visualbert: A simple and performant baseline for vision and language.
\newblock {\em arXiv}, 2019.

\bibitem{li2020unimo}
Wei Li, Can Gao, Guocheng Niu, Xinyan Xiao, Hao Liu, Jiachen Liu, Hua Wu, and
  Haifeng Wang.
\newblock Unimo: Towards unified-modal understanding and generation via
  cross-modal contrastive learning.
\newblock In {\em ACL}, 2021.

\bibitem{li2020oscar}
Xiujun Li, Xi Yin, Chunyuan Li, Pengchuan Zhang, Xiaowei Hu, Lei Zhang, Lijuan
  Wang, Houdong Hu, Li Dong, Furu Wei, et~al.
\newblock Oscar: Object-semantics aligned pre-training for vision-language
  tasks.
\newblock In {\em ECCV}, 2020.

\bibitem{lin2014microsoft}
Tsung-Yi Lin, Michael Maire, Serge Belongie, James Hays, Pietro Perona, Deva
  Ramanan, Piotr Doll{\'a}r, and C~Lawrence Zitnick.
\newblock Microsoft coco: Common objects in context.
\newblock In {\em ECCV}, 2014.

\bibitem{liu2019use}
Yang Liu, Samuel Albanie, Arsha Nagrani, and Andrew Zisserman.
\newblock Use what you have: Video retrieval using representations from
  collaborative experts.
\newblock In {\em BMVC}, 2020.

\bibitem{loshchilov2016sgdr}
Ilya Loshchilov and Frank Hutter.
\newblock Sgdr: Stochastic gradient descent with warm restarts.
\newblock In {\em ICLR}, 2017.

\bibitem{loshchilov2017decoupled}
Ilya Loshchilov and Frank Hutter.
\newblock Decoupled weight decay regularization.
\newblock In {\em ICLR}, 2019.

\bibitem{lu2019vilbert}
Jiasen Lu, Dhruv Batra, Devi Parikh, and Stefan Lee.
\newblock Vilbert: Pretraining task-agnostic visiolinguistic representations
  for vision-and-language tasks.
\newblock In {\em NeurIPS}, 2019.

\bibitem{miech2021thinking}
Antoine Miech, Jean-Baptiste Alayrac, Ivan Laptev, Josef Sivic, and Andrew
  Zisserman.
\newblock Thinking fast and slow: Efficient text-to-visual retrieval with
  transformers.
\newblock In {\em CVPR}, 2021.

\bibitem{nguyen2016ms}
Tri Nguyen, Mir Rosenberg, Xia Song, Jianfeng Gao, Saurabh Tiwary, Rangan
  Majumder, and Li Deng.
\newblock Ms marco: A human generated machine reading comprehension dataset.
\newblock In {\em CoCo@ NIPS}, 2016.

\bibitem{nogueira2019passage}
Rodrigo Nogueira and Kyunghyun Cho.
\newblock Passage re-ranking with bert.
\newblock {\em arXiv}, 2019.

\bibitem{ordonez2011im2text}
Vicente Ordonez, Girish Kulkarni, and Tamara Berg.
\newblock Im2text: Describing images using 1 million captioned photographs.
\newblock {\em NeurIPS}, 2011.

\bibitem{radford2021learning}
Alec Radford, Jong~Wook Kim, Chris Hallacy, Aditya Ramesh, Gabriel Goh,
  Sandhini Agarwal, Girish Sastry, Amanda Askell, Pamela Mishkin, Jack Clark,
  et~al.
\newblock Learning transferable visual models from natural language
  supervision.
\newblock {\em arXiv}, 2021.

\bibitem{sanh2019distilbert}
Victor Sanh, Lysandre Debut, Julien Chaumond, and Thomas Wolf.
\newblock Distilbert, a distilled version of bert: smaller, faster, cheaper and
  lighter.
\newblock In {\em 5th Workshop on Energy Efficient Machine Learning and
  Cognitive Computing, NeurIPS}, 2019.

\bibitem{sharma2018conceptual}
Piyush Sharma, Nan Ding, Sebastian Goodman, and Radu Soricut.
\newblock Conceptual captions: A cleaned, hypernymed, image alt-text dataset
  for automatic image captioning.
\newblock In {\em ACL}, 2018.

\bibitem{sun2019deeply}
Dawei Sun, Anbang Yao, Aojun Zhou, and Hao Zhao.
\newblock Deeply-supervised knowledge synergy.
\newblock In {\em CVPR}, 2019.

\bibitem{sun2021lightningdot}
Siqi Sun, Yen-Chun Chen, Linjie Li, Shuohang Wang, Yuwei Fang, and Jingjing
  Liu.
\newblock Lightningdot: Pre-training visual-semantic embeddings for real-time
  image-text retrieval.
\newblock In {\em NACCL}, 2021.

\bibitem{sun2020mobilebert}
Zhiqing Sun, Hongkun Yu, Xiaodan Song, Renjie Liu, Yiming Yang, and Denny Zhou.
\newblock Mobilebert: a compact task-agnostic bert for resource-limited
  devices.
\newblock In {\em ACL}, 2020.

\bibitem{tan2019lxmert}
Hao Tan and Mohit Bansal.
\newblock Lxmert: Learning cross-modality encoder representations from
  transformers.
\newblock In {\em EMNLP}, 2019.

\bibitem{touvron2021training}
Hugo Touvron, Matthieu Cord, Matthijs Douze, Francisco Massa, Alexandre
  Sablayrolles, and Herv{\'e} J{\'e}gou.
\newblock Training data-efficient image transformers \& distillation through
  attention.
\newblock In {\em ICML}, 2021.

\bibitem{vaswani2017attention}
Ashish Vaswani, Noam Shazeer, Niki Parmar, Jakob Uszkoreit, Llion Jones,
  Aidan~N Gomez, {\L}ukasz Kaiser, and Illia Polosukhin.
\newblock Attention is all you need.
\newblock In {\em NeurIPS}, 2017.

\bibitem{wang2020minivlm}
Jianfeng Wang, Xiaowei Hu, Pengchuan Zhang, Xiujun Li, Lijuan Wang, Lei Zhang,
  Jianfeng Gao, and Zicheng Liu.
\newblock Minivlm: A smaller and faster vision-language model.
\newblock {\em arXiv}, 2020.

\bibitem{xu2016msr}
Jun Xu, Tao Mei, Ting Yao, and Yong Rui.
\newblock Msr-vtt: A large video description dataset for bridging video and
  language.
\newblock In {\em CVPR}, 2016.

\bibitem{young2014image}
Peter Young, Alice Lai, Micah Hodosh, and Julia Hockenmaier.
\newblock From image descriptions to visual denotations: New similarity metrics
  for semantic inference over event descriptions.
\newblock {\em TACL}, 2014.

\bibitem{zhang2021adversarial}
Hang Zhang, Yeyun Gong, Yelong Shen, Jiancheng Lv, Nan Duan, and Weizhu Chen.
\newblock Adversarial retriever-ranker for dense text retrieval.
\newblock {\em arXiv}, 2021.

\bibitem{zhang2021vinvl}
Pengchuan Zhang, Xiujun Li, Xiaowei Hu, Jianwei Yang, Lei Zhang, Lijuan Wang,
  Yejin Choi, and Jianfeng Gao.
\newblock Vinvl: Revisiting visual representations in vision-language models.
\newblock In {\em CVPR}, 2021.

\bibitem{zhang2018deep}
Ying Zhang, Tao Xiang, Timothy~M Hospedales, and Huchuan Lu.
\newblock Deep mutual learning.
\newblock In {\em CVPR}, 2018.

\end{thebibliography}
}

\end{document}